\documentclass{article}

\usepackage{arxiv}

\usepackage[utf8]{inputenc} 
\usepackage[T1]{fontenc}    
\usepackage{hyperref}       
\usepackage{url}            
\usepackage{booktabs}       
\usepackage{amsfonts}       
\usepackage{nicefrac}       
\usepackage{microtype}      
\usepackage{graphicx}
\usepackage{natbib}
\usepackage{doi}
\usepackage{multicol}

\title{3D Patient-specific Modelling and Characterisation of Muscle-Skeletal Districts}

\date{} 					

\author{{\hspace{1mm}Martina~Paccini} \\
	Istituto di Matematica Applicata \\e Tecnologie Informatiche\\ `E. Magenes' CNR\\
	Via de Marini 6 \\ Genova 16149 , Italy\\
	\texttt{martina.paccini@ge.imati.cnr.it} \\
	\And
	{\hspace{1mm}Giuseppe~{Patan{\'e}}} \\
    Istituto di Matematica Applicata \\e Tecnologie Informatiche \\`E. Magenes' CNR\\
	Via de Marini 6 \\ Genova 16149 , Italy\\
	\And
	{\hspace{1mm}Michela~Spagnuolo} \\
    Istituto di Matematica Applicata \\e Tecnologie Informatiche \\`E. Magenes' CNR\\
	Via de Marini 6 \\ Genova 16149 , Italy\\
}

\renewcommand{\headeright}{Technical Report}

\begin{document}
\maketitle

\begin{abstract}
This work addresses the patient-specific characterisation of the morphology and pathologies of muscle-skeletal districts (e.g., wrist, spine) to support diagnostic activities and follow-up exams through the integration of morphological and tissue information. We propose different methods for the integration of morphological information, retrieved from the geometrical analysis of 3D surface models, with tissue information extracted from volume images. For the qualitative and quantitative validation, we will discuss the localisation of bone erosion sites on the wrists to monitor rheumatic diseases and the characterisation of the three functional regions of the spinal vertebrae to study the presence of osteoporotic fractures. The proposed approach supports the quantitative and visual evaluation of possible damages, surgery planning, and early diagnosis or follow-up studies. Finally, our analysis is general enough to be applied to different districts.
\end{abstract}


\section{Introduction}
For decades, shape and medical image analysis have contributed to the development of anatomical visualisation techniques, which remains an active branch of research due to its intrinsic complexity. The variety of anatomical structures and markers that characterise pathological conditions is a key issue. Medical image analysis allows the definition of \emph{patient-specific models}, which are at the core of Computer-Aided Diagnosis (CAD) systems, tailored therapies and interventions. In recent years, CAD has become one of the major research subjects in medical imaging and diagnostic radiology for assisting physicians in the early detection of degenerative pathologies (Sect.~\ref{sec:RELATED-WORK}). A 3D visualisation supports the analysis and rendering of morphological changes, but it cannot provide information on the tissue composition or variation over time. In contrast, focusing only on the image intensities does not provide morphological information as detailed as the 3D model's geometrical analysis, but can highlight the tissue status and differentiate between healthy and pathological tissues. 

\begin{figure}[t]
	\centering
	\includegraphics[width=1\linewidth]{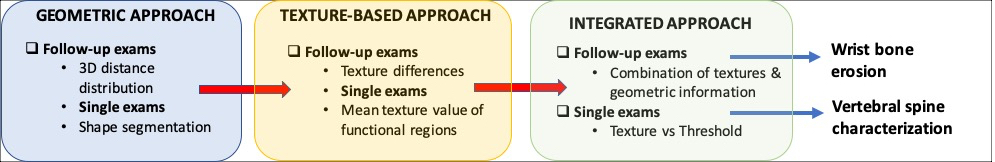}
	\caption{Overview of the proposed approach in terms of its main components and validation scenarios.\label{pipeline}}
\end{figure}

Improving our preliminary results in~\cite{paccini2022combining}, we focus on an effective and interactive representation and analysis of anatomical structures to facilitate the interpretation of the morphology and pathology of the wrist and spine districts. Since both tissue and morphology information is relevant for an accurate analysis, we address the follow-up analysis and patient-specific characterisation of segmented anatomical structures leveraging the integration of geometrical and tissue-related information. We extract information on the morphology of the shape, based on a geometrical analysis of 3D surface models of the bones to identify pathological changes over time or morphological characteristics of the single exam (Sect.~\ref{Sect:Geometry_method}). Moreover, we retrieve the tissue composition in a neighbour of the bone surface. We then propose the integration of the geometric and tissue-related information into a tool for the visualisation and analysis of muscle-skeletal districts and pathologies. In the experimental part (Sect.~\ref{Result}), we discuss the characterisation of (i) erosion sites in the wrist district and (ii) the spine in healthy subjects and patients with osteoporosis. To this end, we use benchmarks of the spine district~\citep{SEKUBOYINA2021102166} and a data set of 3D MRIs of the carpus~\cite{tomatis2015database}. Fig.~\ref{pipeline} shows the structure and main components of the proposed approach.

As the main contribution and novelties (Sect.,~\ref{sec:FUTURE-WORK}), the proposed characterisations are patient-specific and support the early diagnosis of muscle-skeletal pathologies and an accurate evaluation of the patient's status. 3D CT or MR images are visualised together with 3D patient-specific models, thus helping the navigation of the input volume. Our work analyses both the morphology, through the study of the 3D models, and the tissue conditions of muscle-skeletal districts, based on the HU values of the CT images or grey levels of the MRI images. In this way, the analysis and characterisation of the single subject include both aspects, in contrast, previous works focus on either morphology or tissues. In our analysis, we evaluate the whole district, a group of bones, or single bones independently. Finally, our characterisation includes a combination of geometrical and tissue-related parameters and applies to different anatomical districts and image scans.

\section{Related work\label{sec:RELATED-WORK}}
Since our methods will be tested on the analysis of rheumatological pathologies and the evaluation of the spine, we review previous work on the identification of erosion sites on the wrist (Sect.~\ref{sec:EROSION-WRIST}) and the characterisation of the spine (Sect.~\ref{sec:SPINE-CHARACTERISATION}).

\subsection{Erosion sites on the wrist\label{sec:EROSION-WRIST}}
Follow-up exams are commonly used to analyse degenerative diseases in most medical branches. These exams consist of the analysis of the patient's status over time. The importance of follow-up analysis resides in their support of the study of pathology development which, in turn, leads to personalised adjustments to the therapy. In the rheumatological domain, a common sign of degeneration is the presence of bone deformation, usually linked to an erosion process that results in an anomalous change of the tissues' composition and, as a consequence, in the bone morphology. These patients suffer from severe pain and difficulties in the mobility of the interested joints, as a consequence of the anomalous changes of bone relations\cite{palmer2003pain}.

For erosion detection, the MR exam provides a sensitivity higher than CT~\cite{dohn2008detection,scheel2006prospective} and a better differentiation of small erosions when compared to US imaging~\cite{backhaus2002prospective}. Usually, the identification of critical sites, such as erosion or synovitis, has been performed by experts. This type of manual identification is error-prone~\cite{busby2018bias}, time-consuming, and can underestimate the extent of the damage~\cite{figueiredo2018methods}, especially with 3D images. Indeed, the automatic identification of erosion sites and the analysis and quantification of image features through (automated) image processing are largely studied in the field of rheumatic diseases~\cite{gornale2016survey,leung2006automatic,huo2015automatic,langs2008automatic}. With the diffusion of Artificial Intelligence, bone erosion has been identified in end-to-end systems~\cite{murakami2018automatic,rohrbach2019bone}. However, largely reliable and clinically validated data sets for deep learning are not as fairly widespread in radiology as in other medical branches. Since bone erosion not only modifies the tissues' composition but also involves bone morphology, the accurate analysis of the bone shape provides information on the location and extension of erosion sites (if any) on the bone surface.

Thanks to Computer Graphics algorithms (e.g., marching cubes, segmentation), 3D surface models are extracted from 3D images to accurately represent the anatomy of the patient and 3D morphological analysis provides results that could not be obtained by 2D image analysis both in the localisation of morphological degeneration~\cite{cevidanes2010quantification} and surgical planning~\cite{zheng20092d,wong20163d}. Follow-up studies evaluate the evolution of erosion sites on wrist bones exploiting 3D shape analysis on segmented images~\cite{barbieri2015mri}. Analysing parameters extracted from a 3D shape, or a comparison of various shapes, it is possible to support general radiology tasks with a particular focus also on follow-up exams~\cite{banerjee2016semantic,joshi2015registration}. A combination of image texture and shape analysis overcomes the limitations of the single approaches in the analysis of the erosion processes~\cite{paccini2020analysis}.

\subsection{Morphological characterisation of the spine\label{sec:SPINE-CHARACTERISATION}}
The spine is a fundamental element of human anatomy that provides support for our body and organs and guarantees a wide range of mobility. Moreover, the spine deals with load transfer and protects the spinal cord from injuries. All these different tasks can be accomplished thanks to the complex anatomical structure of each element composing the spine and to the organisation of the different tissues in the overall district. Bio-mechanical pathologies, in the short term, can lead to
pain and disability that will worsen in the long term.~\citep{cauley2000risk,SpineAna11:online}.

For the study of the spine, medical imaging supports the evaluation of the status of the different tissues and pathological damage evolution; to this end, CT images are widely used~\citep{bibb2014medical}. The analysed image is a grey-scale image, where the tissue density is represented by shades of grey. The \emph{Hounsfield scale} is a quantitative scale for describing radio density in medical CT and expressing intensity values in \emph{Hounsfield Units} (HU). On the Hounsfield scale, the air is represented by a value of~$-1000$ HU (black on the grey scale) and bone between~$+700$ HU (cancellous bone) to~$+3000$ HU (dense bone) (white on the grey scale). Bones stand out clearly in CT images since they are much denser than soft tissues. Different studies evaluated HU values in CT for the analysis of the physio-pathological status of the subject to help in disease diagnosis~\citep{loffler2020x,zou2019use}; other works identified a correlation between HU values and bone mineral densities~\citep{kim2019hounsfield}. This correlation is important since it allows the use of CT HU values for retrieving different information and quantitative analysis, which, in turn, reduces the need for other types of invasive images, thus reducing the overall amount of ionising radiation for the patient. In clinical practice, the evaluation of the HU values is usually conducted on 2D slices and in the ROI manually identified by experts~\citep{lee2013correlation},\citep{zou2019use},\citep{kim2019hounsfield}, without taking advantage of modern 3D imaging techniques.

Other works focused on the realisation of 3D models to study the spine properties, considering the whole spine or the single vertebra. These models are mainly built for Finite Element analysis~\citep{barron2007generation}, e.g.,~\citep{aroeira2017three,salsabili2019simplifying,campbell2016automated,anitha2020effect} but usually are not patient-specific. Indeed, when it comes to assigning material properties of the tissue, these works rely on the state-of-the-art instead of on the single subject case. A common technique for the development of vertebral models includes statistical shape models~\citep{castro2015statistical,campbell2016automated} or parametric model~\citep{vstern2011parametric}, which adapt to the subject-specific case with a certain level of accuracy, but usually require the construction of a prior reference.

For a deeper evaluation of the spine biomechanics, various studies focused on alignment analysis~\citep{laouissat2018classification,yeh2021deep,roussouly2011sagittal} and the study of the vertebral spine or single vertebra morphology~\citep{keller2005influence,lois20193d,casciaro2007automatic,shaw2015characterization}. The main drawback is the use of 2D images, which provide a limited visualisation of the spine. Even though rendering techniques have been developed in recent years, the common practice still considers individual slices. However, recent studies have shown that the 3D information of the vertebral spine can make a difference in early diagnosis or subject evaluation~\citep{labelle2011seeing}. In contrast to 2D image analysis, the use of 3D information allows the inclusion of asymmetries in the evaluations~\citep{barron2007generation}, improves the estimation of curvatures~\citep{lois20193d}, and encodes information on the vertebral disc status~\citep{fazzalari2001intervertebral,lois20193d}. Most of these studies start from CT images, which have the advantage to preserve the real distances between vertebral bodies and the real curvature information~\citep{barron2007generation}. However, previous work on 3D modelling or imaging usually concentrates only on a specific region or motion segment and not on the complete vertebral spine~\citep{barron2007generation}, as in~\citep{lois20193d}.

\section{Geometric and texture-based features of anatomical districts\label{Sect:Geometry_method}}
We now discuss the extraction of geometric (Sect.~\ref{geom_single}) and texture-based (Sect.~\ref{SECT:texture_methods}) features, which are then combined in a unified approach (Sect.~\ref{SECT:integratedMethods}).
\begin{figure}[t]
	\centering
	\begin{tabular}{ccccc}
		(a)\includegraphics[height=0.17\linewidth]{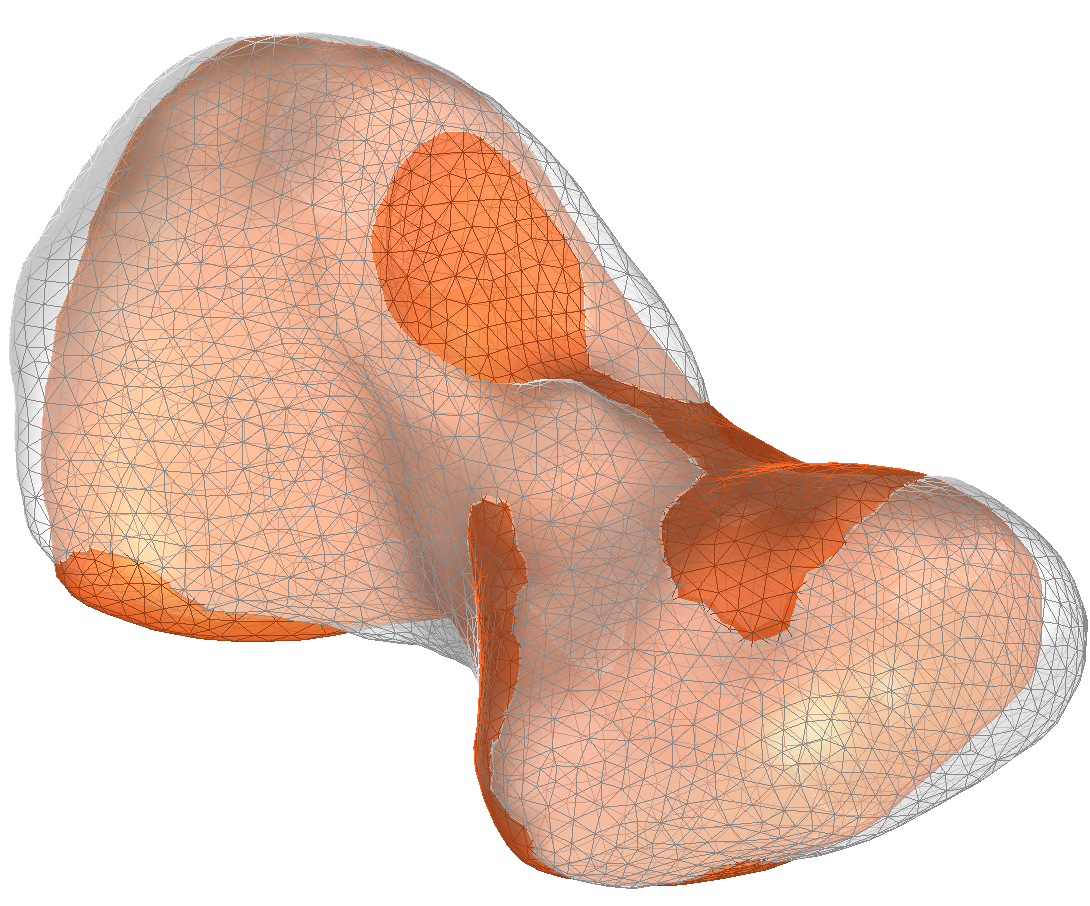}&
		(b)\includegraphics[height=0.17\linewidth]{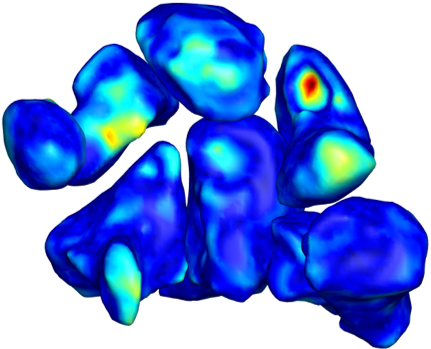}
		&(c)\includegraphics[height=0.17\linewidth]{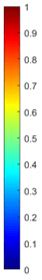}&\includegraphics[height=0.17\linewidth]{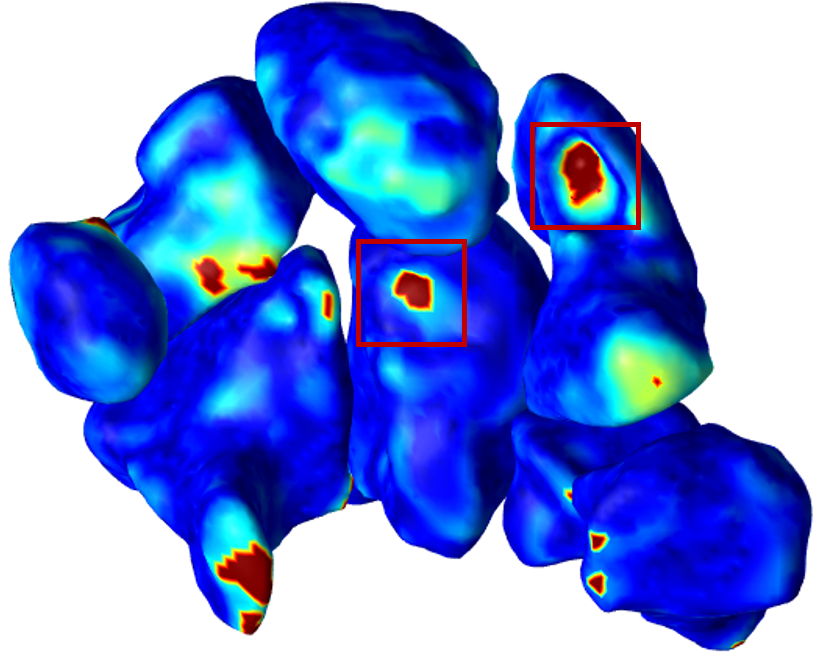}
	\end{tabular}
	\caption{(a) Registration result between the 3D surface model of a single bone at baseline time and follow-up time. (b-c) Geometry-based features extraction: distribution of local distances between the bones at baseline time and follow-up time. Red rectangles highlight pathological deformation localised by experts. The original value in millimetres has been normalised between 0 and 1.\label{geometry features}}
\end{figure}
\subsection{Geometric features of single and follow-up exams \label{geom_single}}
For a \emph{single exam}, we considered the geometry of the 3D surface model of the bone at a specific time and focused on shape segmentation, by identifying and labelling the different regions of interest or \emph{functional regions} that characterise the bone according to anatomical studies. The shape segmentation consists of the association of each vertex of the bone surface to a particular functional region. Then, we consider the geometrical features that discriminate the functional region where each vertex belongs. Since the segmentation that we implemented is specific to the district, we refer the reader to the experimental section for more detail (Sect.~\ref{Result}). 

To evaluate the variation of geometric features of bones in \emph{follow-up exams}, we compare the 3D surfaces of the bones at baseline time (${t_{1}}$ (i.e., first evaluation)) with the one at follow-up time (${t_{2}}$). The analysis is carried out by rigidly co-registering the surface models belonging to the two exams and evaluating local distances to identify shape changes over time (Fig.~\ref{geometry features}a). Then, we compute the Hausdorff distance between the two co-registered districts to identify the bones with pathological deformation. Calling~$\mathbf{X}_{1}$ the 3D bone surface at~$t_{1}$ and~$\mathbf{X}_{2}$ the coregistered 3D surface at~$t_{2}$, we identify eroded bones by their Hausdorff distance: \mbox{$d(\mathbf{X}_{1},\mathbf{X}_{2}):=\max\{d_{\mathbf{X}_{1}}(X_{2}),d_{\mathbf{X}_{2}}(X_{1})\}$}, where \mbox{$ d_{\mathbf{X}_{1}}\left(\mathbf{X}_{2}\right):=\max _{\mathbf{x}\in \mathbf{X}_{1}}\left\{ \min _{\mathbf{y}\in \mathbf{X}_{2}}\left\{ \left\|\mathbf{x}-\mathbf{y}\right\| _{2}\right\} \right\}$}. We then evaluate the local distribution of the minimum distance of each vertex of the surface at~$t_{2}$ from the vertices of the surface at~$t_{1}$. The distance distribution is normalised to~$[0,1]$ to be comparable with the results obtained with the texture approach. The regions of the bone, where the morphology has changed the most, present a higher value of distance distribution (close to 1), thus highlighting possible erosion processes (Fig.~\ref{geometry features}, (b-c)).
\subsection{Texture-based features of single and follow-up exams\label{SECT:texture_methods}}
\begin{figure}[t]
	\centering
	\begin{tabular}{ccc}
		\includegraphics[width=0.30\linewidth]{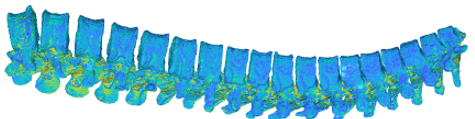}
		&\includegraphics[width=0.30\linewidth]{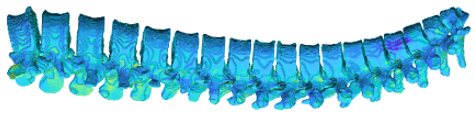}
		&\includegraphics[width=0.30\linewidth]{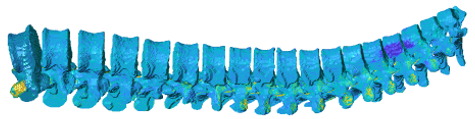}\\
		(a) &(b) &(c)
	\end{tabular}
	\caption{\label{fig:mappindResult} Texture extraction through the application of the mapping method to vertebral surface model in the three different criteria: (a) internal, (b) Euclidean and (c) external. The difference in texture reflects the differences in tissue composition.}
\end{figure}

Following~\cite{paccini2020analysis}, we generate a texture for the 3D bone surface extracted from an input image of a \emph{single exam}. The segmented 3D surface is a triangle mesh, and the volume image is composed of a series of 2D slices and is represented as a voxel grid. The image is loaded into a 3D grid, whose elements have the same dimension as the image voxels. In this way, every grid cell has its 3D coordinates and each grid cell is associated with a voxel and its grey level. This data structure allows the navigation of the volume through the surface to find the correct correspondences between the volume image and the segmented 3D surface. For the grey value mapping, three different criteria are defined to analyse the grey values in the proximity of the surface. Such criteria depend on the method chosen to identify the correspondences between the surface vertices and the volume voxels. In the \emph{Euclidean mapping}, the surface vertex~$\mathbf{p}$ gets the grey level of the voxel closest to~$\mathbf{p}$. In the \emph{internal mapping}, the closest voxels are searched only inside the surface, i.e., inside the object's volume. In the \emph{external mapping}, the closest voxels are searched only outside the surface, i.e., outside the object's volume. Once every vertex of the surface has been associated with a voxel, the same grey level is associated with the correspondent voxel. The result of the grey-level mapping is a textured surface, where each vertex coordinate is associated with its specific colour, representative of the information contained in the image (Fig.~\ref{fig:mappindResult}). 
This result is satisfactory if we are evaluating a \emph{single exam} since we can directly consider the grey-level characterising tissue in the neighbourhood of the bone surface from the texture. The mean value of texture in a particular region of the 3D surface model will be considered as the texture features in the following integration section. To consider all the volume surrounding the bone surface, the mean value of the grey levels can be computed for all the different mapping methods: internal, external and Euclidean. In this way, the analysis will be more exhaustive. 

For \emph{follow-up exams}, we compare the tissue evolution over time. For this reason, we now describe how to evaluate the tissue changes from the textured 3D surface. After co-registering the two surfaces, the follow-up 3D model is textured with the grey levels of both the image at baseline time ($t_{1}$) and at follow-up time ($t_{2}$), exploiting the grey-levels mapping. Mapping the volume grey levels at time~$t_{1}$ on the surface at time~$t_{2}$, the follow-up 3D model shows either the volume situation at time~$t_{1}$ and at time~$t_{2}$. We focus on external mapping, which explores the volume toward the external normal directions from the surface vertices since erosion typically results from a degeneration of articular regions. The variations in texture are highlighted by computing the difference between the grey levels mapped from the baseline image onto the follow-up surface and the grey levels mapped from the follow-up image onto the follow-up surface. A higher value of difference indicates a region where the tissue has been substantially damaged, probably due to an erosion process. The colour map represents the tissue modification directly on the 3D surface.

\subsection{Integrated approach\label{SECT:integratedMethods}}
We now describe the integration of geometrical and texture-based features considered to obtain a unique and complete evaluation of both the follow-up and the single exams. In the case of a \emph{single exam}, we have a 3D surface textured with the original volume grey levels, meaning that each vertex of the triangulation is associated with the grey level values extracted from the image. As a geometrical feature, we consider the geometrical property that classifies each vertex in its functional region, as texture information we consider the mean grey level of each functional region. This means that we can associate to the geometrical property, the relative mean grey level value of the region that it identifies on the surface, to visualise both properties in a unique analysis. We then compare how this association distributes over all the bones of the district of different subjects.  

For the analysis of \emph{follow-ups}, we present two approaches that integrate the information extracted from the geometry and texture information. The distance distribution (geometrical information) is associated with each bone vertex at follow-up time, and express the value of the distance between the bone at follow-up with the corresponding bone at baseline. From the texture analysis, each vertex of the follow-up district is associated with the value of the difference between the image grey levels at baseline and follow-up. In this way, each vertex composing the segmented surface in the follow-up is associated with a feature vector, composed of geometrical and texture information. In particular, the texture information presents intensities from -1 to 1, since they are obtained from the difference of grey levels between 0 and 1. A value near 1, indicates a major change in the intensity of the image voxels, i.e., a radical change in tissue composition over time. A value of texture near -1, might indicate a healing process, where the extension of an existing erosion has been reduced due to a specific therapy~\cite{barbieri2015mri}.

The value of the distance, instead, varies from 0 to 1 depending on the entity of the geometric deformation. The first option is to multiply the geometry and the texture values for each vertex. In this way, if a variation in geometry is associated with a high variation in texture (value near 1), then the result has a higher total value. If a geometry change is close to zero, then the influence of the geometry has an impact on the total value lower than the grey levels in the texture. In this way, the pathological changer will be highlighted only if texture and geometry, show a major variation, thus confirming that the bone has developed both a change in morphology and tissue composition. As a second option, we consider a linear combination of geometric and texture information. Calling~$d_2$ the distribution of the distances obtained by the geometry analysis and~$d_1$ the distribution of the texture changes in the grey-levels analysis, the combined distribution is defined~$d_1+\epsilon d_2$ with~$\epsilon>0$. Varying the value of~$\epsilon$, it is possible to emphasise geometric or texture information.
\section{Results\label{Result}}
As experimental tests, we discuss the evaluation of erosion sites as proof for the follow-up exams integrated approach (Sect.~\ref{wristResult}), and the spine characterisation as proof for the single exam integrated evaluation (Sect.~\ref{spineResult}).

\subsection{Follow-up evaluation: erosion sites' identification for the wrist district\label{wristResult}}

The data set used for the analysis of the carpal district is composed of the 3D MRIs of the carpus associated with their segmentation. The MRIs are obtained from a 0.2T extremity-dedicated machine (Esaote-Artoscan) using 3D T1-weighted sequences with reconstruction on the axial and sagittal plane. Voxel intensities have been normalised between 0 and 1, being zero black and one white. The clinical experts performed the semi-automatic segmentation of the bones with a region-growing method~\cite{tomatis2015database} supported by the CAD system RheumaSCORE~\cite{parascandolo2014computer}. Since the segmentation has been performed by medical doctors, segmented surfaces are considered ground truth without applying further processing, such as smoothing and re-meshing. The subjects suffer from degenerative rheumatoid illnesses at different stages. The carpus is one of the most indicated for diagnosing and monitoring rheumatic diseases since it is one of the first to show symptoms and damages linked to the pathology. For our analysis, we are interested in the 5 subjects that underwent follow-up exams, to analyse the pathology progression. For each of those subjects, 2 exams are present: one at baseline time ($t_{1})$ and one at follow-up time ($t_{2}$). The physicians or radiologists can visualise either the single bone or the overall district based on his/her necessity, but the changes in each bone over time are analysed separately. We included also the hand bones and the distal segment of the forearm bones, thus, we compare how 125 bones changed over time.

\begin{figure*}[t]
	\centering
	\begin{tabular}{ccccc}
		(a)\includegraphics[height=0.20\linewidth]{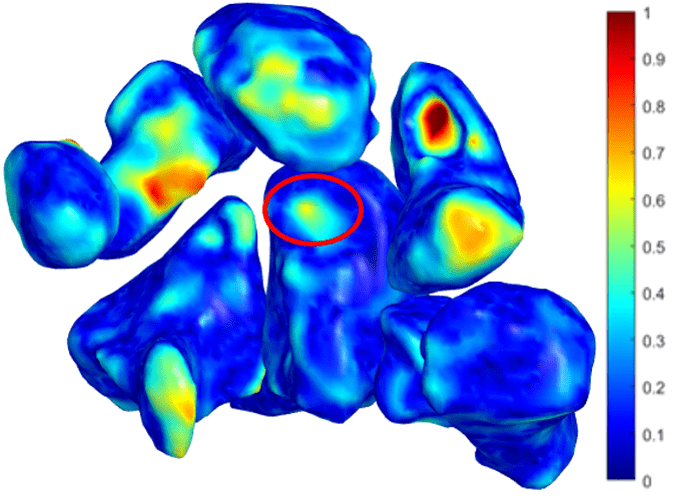}&
		(b)\includegraphics[height=0.20\linewidth]{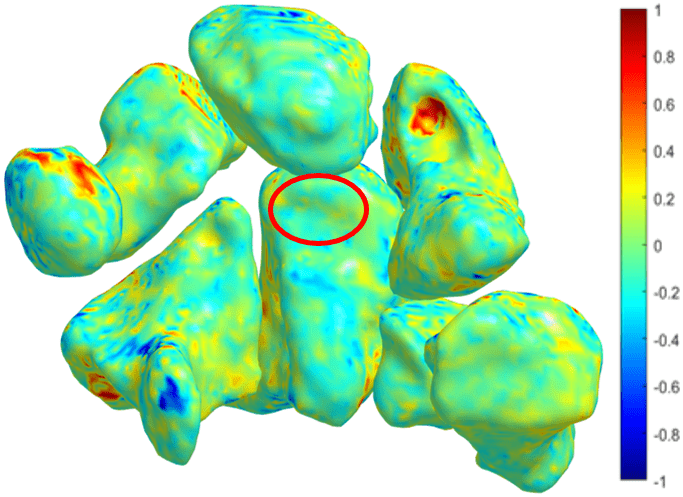}&
		(c)\includegraphics[height=0.20\linewidth]{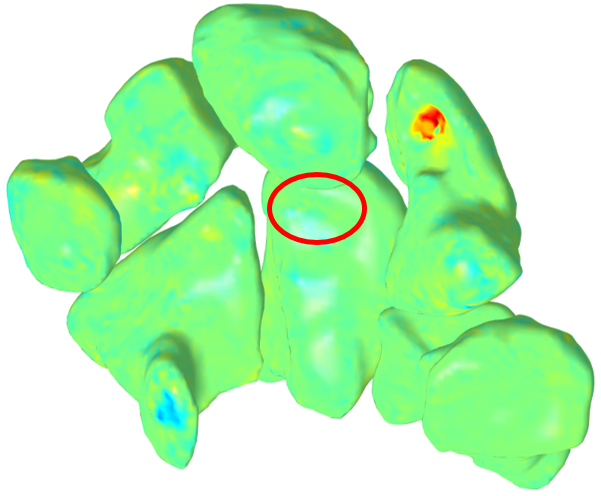}
	\end{tabular}
	\caption{(a) Comparison of the geometry features, (b) texture features, and (c) their integration through the multiplication of the geometric and texture intensities, on a follow-up structure. The colour map presents the higher intensity values of the distribution as warm colours, where areas with the highest values are red and cold colours represent the smaller values in the distribution. The red circle indicates the small erosion identified by the geometry-based approach but not by the integration method. The original values in millimetres and image intensity are normalised before the integration\label{comparative result}}
\end{figure*}

The integrated method merges the information provided by the geometrical analysis and texture analysis. Considering only the geometrical or the texture information, the result could present false positives, i.e., areas identified as erosion but that are healthy~\cite{paccini2022combining}. In Fig.~\ref{comparative result}(c), false positives are not present in the results of the integrated approach, where only the areas with erosion show higher integration values. The overall result is more homogeneous, thus indicating that geometry and texture information compensate for each other, discarding misleading classifications due to low image resolution or registration inaccuracies. Nevertheless, the integration result could miss some newly developed and small erosions. The red circles in Fig.~\ref{comparative result} indicate the small erosion correctly localised by the geometrical information, but missed by texture analysis. This suggests that the erosion could be at its beginning, implying a minor change in geometry and a small change in the cortical bone tissue, which could be still partly intact. 
\begin{figure*}[t]
	\centering
	\begin{tabular}{ccc}
		(a)\includegraphics[height=0.20\linewidth]{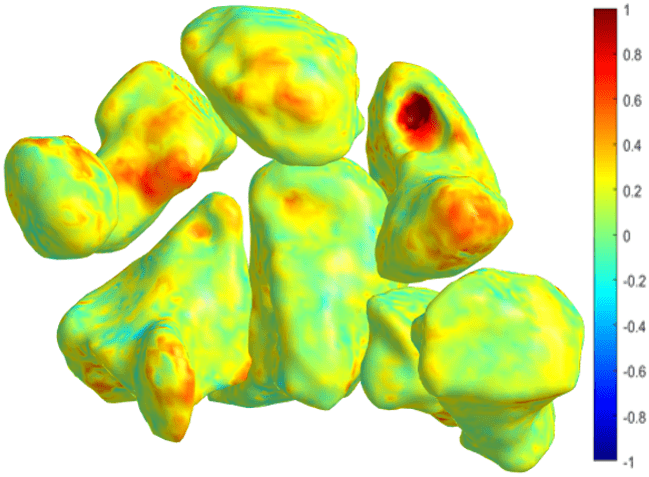}&
		(b)\includegraphics[height=0.20\linewidth]{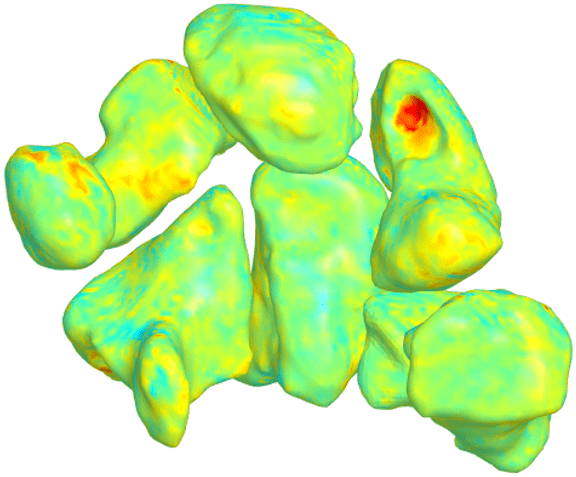}&
		(c)\includegraphics[height=0.20\linewidth]{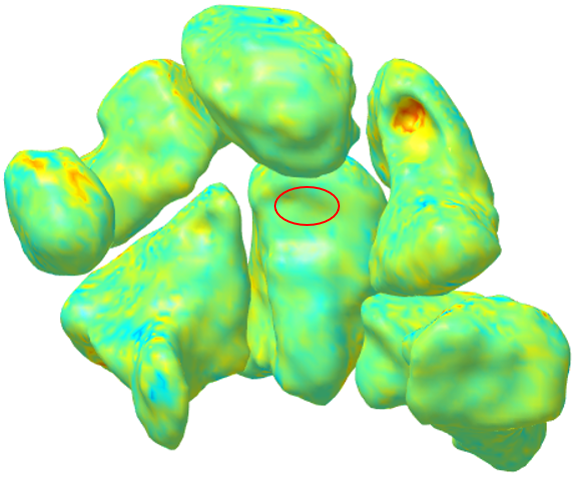}
	\end{tabular}
	\caption{Linear combination integration method varying~$\epsilon$:~$d_{1}+\epsilon d_{2}$ where~$d_{1}$ is the texture distribution and~$d_{2}$ is the geometry distribution; (a) for~$\epsilon=1$, that is the resulting distribution is given by~$d_{1}+d_{2}$  (b) for~$\epsilon=0.5$, the resulting distribution is given by~$d_{1}+0.5d_{2}$, (c) for~$\epsilon=0.2$, that is the resulting distribution is given by~$d_{1}+0.2d_{2}$ and the red circle highlight the false negative. The colour map presents as warm colours the higher intensity values of the distribution, where red areas have the highest results; cold colours represent the smaller values in the distribution.\label{linear combination result}}
\end{figure*}
The integration approach, which considers the linear combination of texture and geometry, is more flexible compared to the multiplication approach. Varying the value of~$\epsilon$ it is possible to tune the analysis concerning the type of search that the physician is carrying on. Fig.~\ref{linear combination result}, shows three different integration results according to the variation of the~$\epsilon$ value. Reducing the relevance of the geometric information ($\epsilon=0.5$,~$\epsilon=0.2$), the resulting values lead to a more homogeneous result. This means that the false positive areas decrease in number along with the reduction of the geometric information. However, an excessive reduction of the geometrical information brings also the presence of false negatives, that is the misclassification of erosion regions as healthy areas. Overall, the linear combination approach allows us to adjust the focus of the search and reach the desired trade-off between false negatives and false positives.
The possibility to manually change the parameter~$\epsilon$ allows physicians and rheumatologists to adjust the visualisation accordingly to their knowledge of the patient. If the physician analyses a patient at the early stages of pathology, then higher values of~$\epsilon$ will help to localise an early erosion. In the case of a patient with an existing localised erosion, lower values of the parameter~$\epsilon$ could highlight the progression of the erosion itself. Moreover, the manual setting of~$\epsilon$ provides support for the quantitative standard metrics used for erosion in rheumatic diseases as the OMERACT rheumatoid arthritis MRI scoring system (RAMRIS) and the EULAR-OMERACT rheumatoid arthritis MRI reference image atlas~\cite{ostergaard2005introduction}. These scoring systems evaluate each bone erosion separately on a scale~$0-10$, based on the proportion of eroded bone compared to the assessed bone volume, judged on all available images. With a manual setting of~$\epsilon$, given the score, it could be possible to assess the level of erosion with higher precision, showing interactively the result of the evaluation on the 3D model. 
The integration methods do not need previous knowledge of the analysed structure, or the computation of mean shapes or atlases to perform the follow-up analysis~\cite{de2004multi}. Finally, the integrated method is quite general and can be applied to different anatomical structures.
\begin{figure}[t]
	\centering
	\begin{tabular}{cc}
		(a)\includegraphics[width=0.35\textwidth]{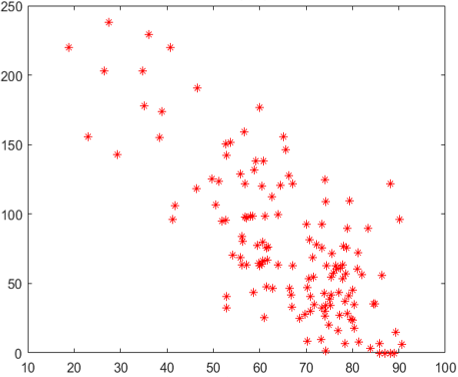}& 
		(b)\includegraphics[height=0.30\textwidth]{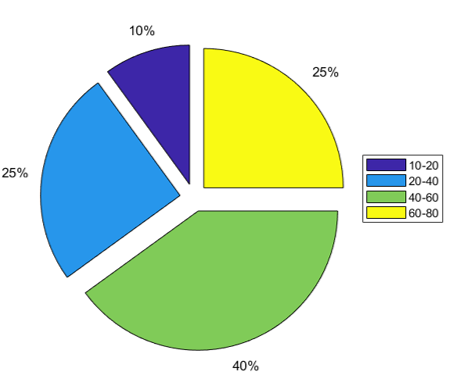}
	\end{tabular}
	\caption{\label{fig:densityAndAge}(a) (Bone Mineral Density) BMD behaviour in function of subject age;~$x$-axis presents subject to age,~$y$-axis BMD values. (b) Age distribution of the considered subjects.}
\end{figure}
\begin{figure}[t]
	\centering
	\begin{tabular}{cc}
		\includegraphics[height=0.20\textwidth]{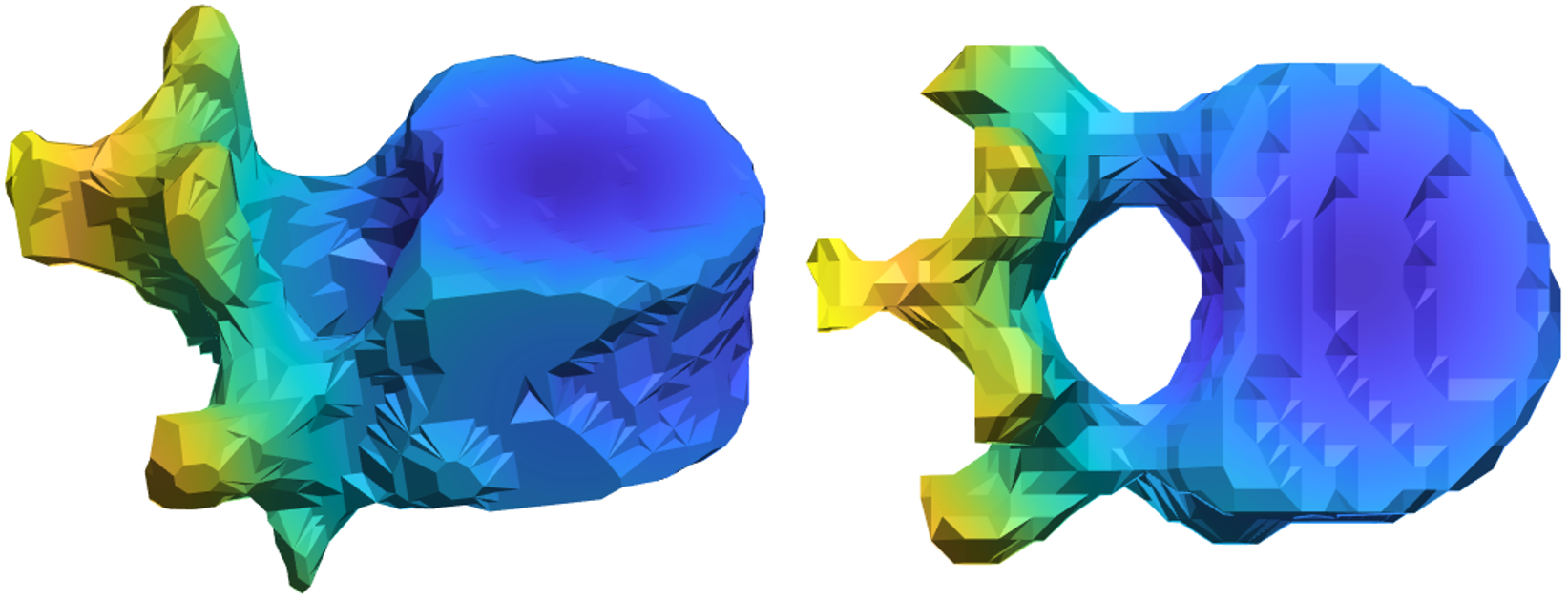}
		&\includegraphics[height=0.20\linewidth]{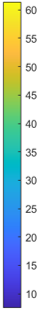}
	\end{tabular}
	\caption{\label{fig:distanceDistribution} Distance distribution of a vertebral bone. Each vertex is associated with the value of its distance from the vertebral body centroid. The unit measure of the colour map is millimetres.}
\end{figure}
\subsection{Single exam evaluation: vertebral spine characterisation\label{spineResult}}
For the analysis of a \emph{single exam}, as the main example, we consider the spine district.
The CT data used for the spine characterisation are a sub-part of the \emph{Large Scale Vertebrae Segmentation challenge} (VerSe), which provides a common benchmark for current and future spine-processing algorithms~\citep{SEKUBOYINA2021102166}. The data are multi-site acquired using multiple CT scanners, present a variety of FoVs (including cervical, thoracolumbar and cervical-thoracolumbar scans), and a mix of sagittal and isotropic reformations, and cases with vertebral fractures, metallic implants, and foreign materials~\citep{SEKUBOYINA2021102166}. The annotations in the data set are twofold and present 3D coordinate locations of the vertebral centroids and voxel-level labels as segmentation masks. Twenty-six vertebrae (C1 to L5, and the transitional T13 and L6) are annotated with labels from 1 to 24, along with labels 25 and 28 for L6 and T13, respectively. This CT data set contains 160 image series of 141 patients including segmentation masks of 1725 fully visualised vertebrae~\citep{doi:10.1148/ryai.2020190138}. The metadata includes annotation of vertebral fractures obtained through Gentant's method, the indication of foreign material, and the measurement of lumbar Bone Mineral Density (BMD) per patient age (Fig.~\ref{fig:densityAndAge}(a)). For our purposes, we used only one type of CT scan, to be able to compare the results obtained in the tissue evaluation. Fig.~\ref{fig:densityAndAge}(b) shows the age distribution of the subjects considered in our work. 
\begin{figure*}[t]
	\centering
	\begin{tabular}{c|c}
		\includegraphics[height=0.25\textwidth]{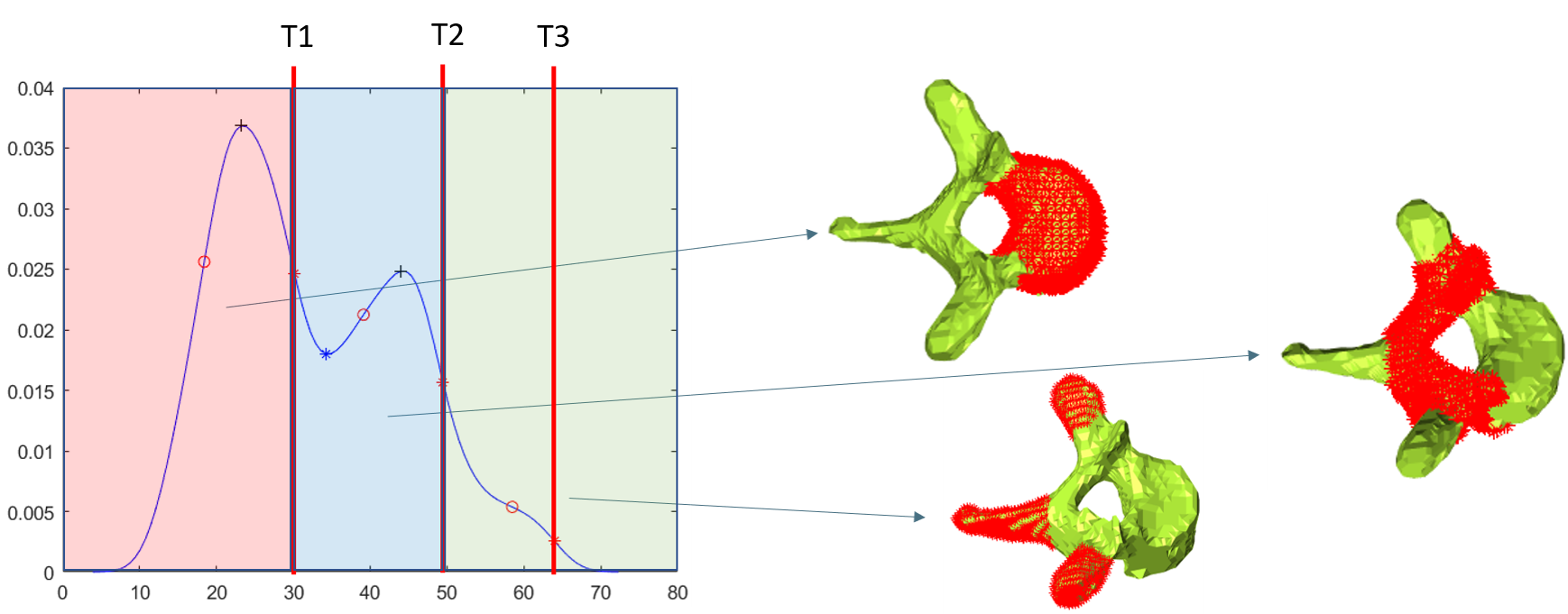}&
	\end{tabular}
	\caption{\label{fig:probabilityDensity}(a) Probability density curve of a vertebra and resulting segmentation associated with the inflexion points. Distances from the centroid values ($x$-axis) and probability density value ($y$-axis). The distance values corresponding to an inflexion point on the curve are considered as thresholds for the vertebral shape segmentation:~$T_{1}$ and~$T_{2}$ create the distinction between the three functional parts. Where~$T_{1}$ is considered the reference value of distance for the vertebral body,~$T_{2}$ for the vertebral arch and~$T_{3}$ for the spinous and transverse processes region.}
\end{figure*}
\begin{figure}[t]
\centering
\begin{tabular}{cc}
	(a)\includegraphics[height=0.13\linewidth]{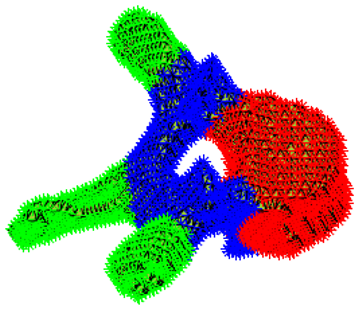}
	(b)\includegraphics[height=0.11\textwidth]{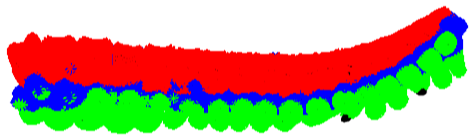}&
	(c)\includegraphics[height=0.11\textwidth]{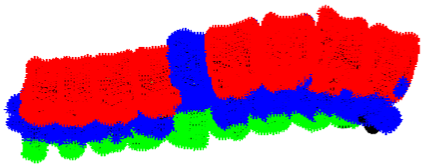}
\end{tabular}
\caption{\label{fig:segmentationSpine} Spine segmentation result: (a) single vertebra shape segmentation, (b) in a healthy subject and (c) in a pathological subject.}
\end{figure}

The segmentation of the vertebral bones localises, the three main functional components of each vertebra (i.e., vertebral body, vertebral arch and transverse processes) from a patient-specific perspective. The first step is the computation of the distance distribution of each vertex composing the vertebral surface from the centroid of the vertebral body. The idea is to grow a sphere centred in the centroid and to assign to each vertex of the bone a value of distance equal to the radius of the first sphere containing it. For efficiency, the sphere growth is executed by computing the Euclidean distance \mbox{$d(\mathbf{p}_{i},\mathbf{c}):=\|\mathbf{p}_{i}-\mathbf{c}\|_{2}$} of each vertex~$\mathbf{p}_{i}$ from the vertebral body centroid ~$\mathbf{c}$ (Fig.~\ref{fig:distanceDistribution}).
\begin{figure}[t]
\centering
\begin{tabular}{cc}
	(a)\includegraphics[height=0.25 \linewidth]{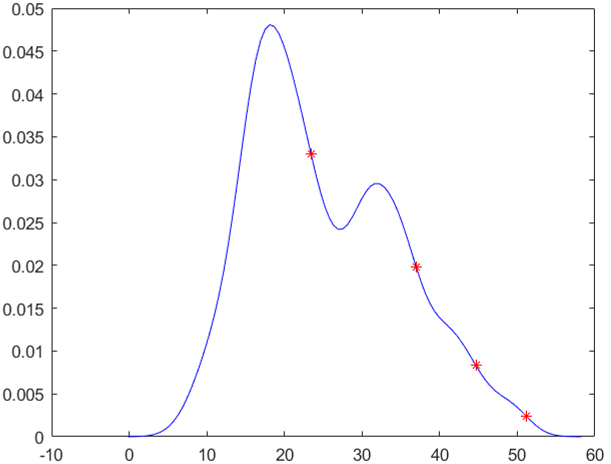}&
	(b)\includegraphics[height=0.25 \linewidth]{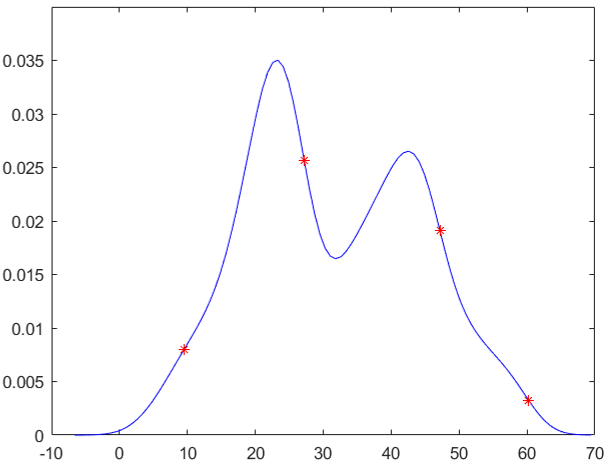}
\end{tabular}
\caption{\label{fig:probDistrPatient} Difference between curve morphology and inflexion point locations in (a) healthy subjects and (b) in pathological subjects.}
\end{figure}

Then, we compute the \emph{probability density} from the distribution of distances associated with the vertices of each bone. In this way, we transform 3D information, such as the distance of the vertex from the centroid, into 1D information. In different subjects, the same vertebra presents a similar behaviour of the probability density curve. Thus, we leverage the inflexion points of the probability density curve as distance thresholds. Indeed, the vertex that has a distance lower than the first inflexion point ($T_{1}$) is considered a part of the vertebral body. The vertex whose distance belongs to the range between the first and the second inflexion point ($T_{2}$) is considered a part of the vertebral arch. All the vertices that have a distance higher than the second inflexion point are classified as part of the transverse or spinous processes and the distance value considered as a threshold is the third inflexion point ($T_{3}$) (Fig.~\ref{fig:probabilityDensity}). Once the vertebral bone has been segmented and labelled in its functional sub-parts, we consider, as a geometrical parameter, the distance values corresponding to the inflexion points (i.e.,~$T_{1}$,~$T_{2}$,~$T_{3}$). Indeed, the threshold used for vertebral segmentation is the geometrical property that permits the classification of the surface vertices. 

Fig.~\ref{fig:segmentationSpine}(a) shows the results of the thresholding of the distance distribution which leads to the shape segmentation. The red part of the surface represents the vertebral body, the blue one the vertebral arch and the green one the transverse and spinous processes. Applying the segmentation to all the vertebrae represented in the image, we obtain the segmentation of the whole vertebral spine (Fig.~\ref{fig:segmentationSpine}(b)). On a pathological subject, the shape and morphology of the vertebra can change drastically as a result of different processes that damage the tissue, thus the 3D surface models highlight such morphological changes. Applying our segmentation to a pathological case that presented a vertebral fracture due to osteoporosis, the fractured vertebra was identified among the others by a clear change in the segmentation result (Fig.~\ref{fig:segmentationSpine}(c)). Indeed, a change in the distance from the vertebral centroid distribution, in turn, modifies the vertebra's probability distribution curve as well as the position of the inflexion points. Fig.~\ref{fig:probDistrPatient} highlights how the inflexion points' location on the probability distribution has changed due to the osteoporosis processes that brought the vertebral fracture. 
\begin{figure}[t]
\centering
\begin{tabular}{ccc}
	\includegraphics[width=0.3 \linewidth]{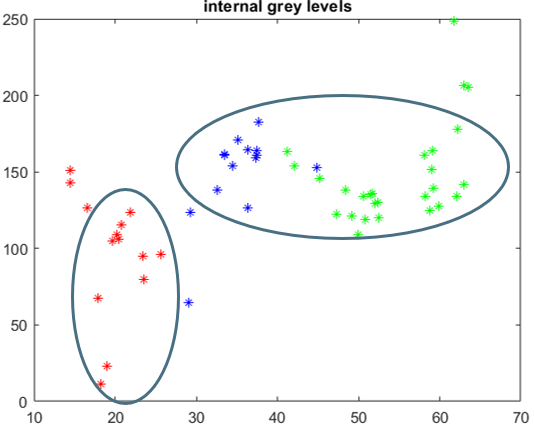}&
	\includegraphics[width=0.3 \linewidth]{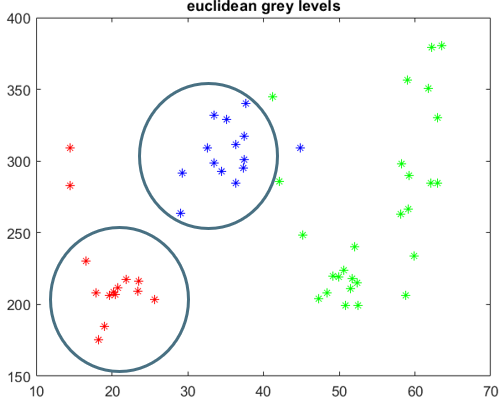}&
	\includegraphics[width=0.3 \linewidth]{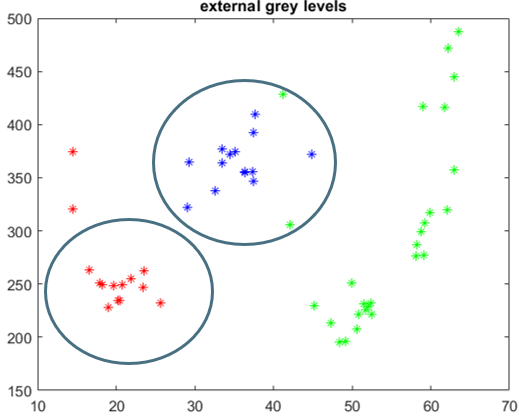}
\end{tabular}
\caption{\label{fig:comparisonResult}Comparison between tissue and geometrical information in the surface neighbourhood.The~$x$-axis shows the thresholds distance values from the centroid; the~$y$-axis the mean HU values of the relative functional area localised by the thresholding, i.e. red points refer to the vertebral body, blue to vertebral arch and green to the spinous and transverse process.}
\end{figure}

The application of the grey-level mapping algorithm to the vertebral surface model produces a 3D textured representation of the spine, where the texture corresponds to the HU values of the original 3D CT. Depending on the mapping criteria we can investigate the tissues inside, across or outside the vertebral surface. As a texture feature, we evaluate the mean HU values of the vertebra's functional region. At this point, we link the texture information obtained through the texture mapping method to the geometric geometrical parameter that permitted the segmentation in functional regions. Thus, we correlate the geometric thresholds localising the vertebral region (i.e.,~$T_{1}$,~$T_{2}$, and~$T_{3}$ in Fig.~\ref{fig:probabilityDensity} (a)), with the mean HU value of the region in all three mapping method criteria. In this way, we characterise the district considering both tissue and geometric information and we evaluate which of the mapping criteria is best suited for this characterisation.

Fig.~\ref{fig:comparisonResult} shows the results obtained in the comparison of geometrical and tissue information. In each different mapping criterion, the graphs show various clusters. In the internal mapping case, the clusters could be associated with the varying percentage of cortical bone in the different vertebral components. Indeed, the body is composed mainly of cancellous bone, a spongy type of bony tissue. This cancellous bone is covered by a thin coating of cortical bone (or compact bone), which is a hard and dense type of bony tissue. The vertebral arch and processes have thicker coverings of cortical bone. The clusters identified in the external mapping results can be related to the tissue surrounding the vertebra from the outside. The vertebral body confines by cartilaginous tissues (intervertebral disc) while the other parts of the vertebral border with ligaments and tendons (e.g. vertebral arch with the foramen). The slightly different values of HU between the Euclidean and external mapping results can be explained by the ligament insertion distribution on the vertebral arch and body. Indeed, the bond between two vertebral bodies is reinforced anteriorly by the anterior longitudinal ligament and posteriorly by the posterior longitudinal ligament, known also as the anterior and posterior spinal ligaments, respectively.
\begin{figure}[t]
\centering
\begin{tabular}{ccc}
	\includegraphics[width=0.3 \linewidth]{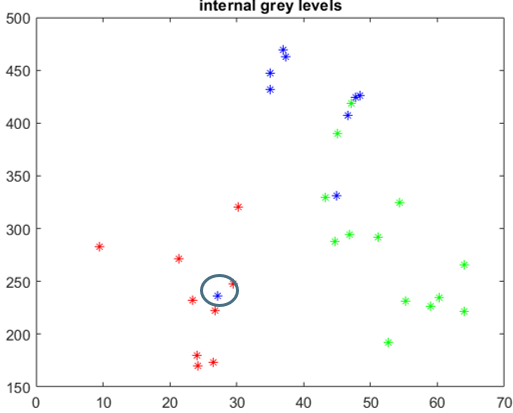}&
	\includegraphics[width=0.3 \linewidth]{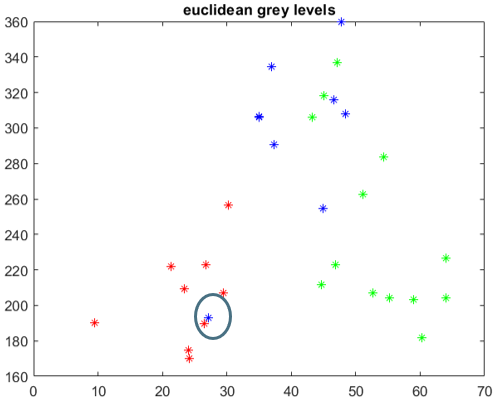}&
	\includegraphics[width=0.3 \linewidth]{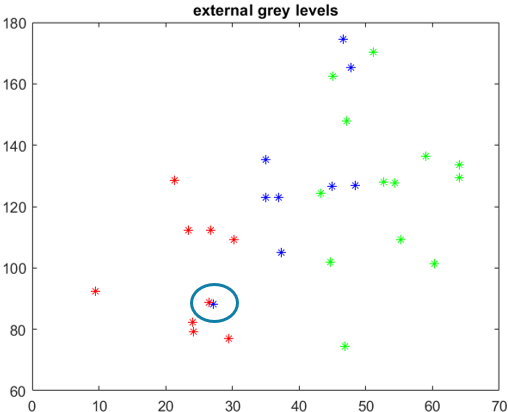}
\end{tabular}
\caption{\label{fig:comparisonResultPatient}Comparison between tissue and geometrical information in the surface neighbourhood in presence of a pathological case. The~$x$-axis shows the thresholds distance values from the centroid; the~$y$-axis the mean HU values of the relative functional area localised by the thresholding, i.e. red points refer to the vertebral body, blue to vertebral arch and green to the spinous and transverse process.}
\end{figure}

In the remaining parts of the vertebrae, there is no intervertebral disc but, in addition to the facet joints that connect adjacent vertebral arches, there are a few strong and bio-mechanically important ligaments that run between adjacent vertebral arches. These ligaments fuse with the anterior and posterior parts of the circumference of the disc. Indeed, Euclidean mapping, which considers the voxel crossed by the surface, is influenced by both the aspects just exposed. In the case of healthy subjects, the distribution of the points in the combined graph could represent, from a clinical point of view, the patient-specific insight into the healthiness of the patient. Comparing the result with the combined analysis of a pathological subject (Fig.~\ref{fig:comparisonResultPatient}), the point in the graph related to the vertebral body of the patient (which is the functional part mostly affected in this case) is an outlier of the results of the healthy subject. Both the graphs related to the internal and external mapping can differentiate between healthy and pathological subjects; in fact, once the osteoporosis processes damage and erode the bone tissues, the cartilaginous tissues are involved and suffer as well. This means that the tissue changes are highlighted by exploring the volume inside the surface and analysing the volume in the region outside the bone surface. Thus, these representations of combined information can highlight pathological cases or vertebrae that differentiate from the classical health distribution, making it a useful tool for personalised and efficient analysis.

\section{Conclusions and future work\label{sec:FUTURE-WORK}}
For both the applications of the method proposed (follow-up exams evaluation on the wrist and single exam characterisation on the spine), the integrated visualisation is coupled with existing traditional techniques to deepen the analysis by providing comprehensive information on both morphological and tissue degeneration in a unique augmented representation. Indeed, the improvement of our approach resides in the possibility to integrate geometry- and texture-based representations of a morphological district and the underlying pathology in an interactive way to better support the analysis of the patient status by multidisciplinary experts. Moreover, the methods are general enough to be applied to different imaging techniques, as proved by the different data sets used and described in the experimental section (Sect.~\ref{Result}).

For the carpal district, our tool could help speed up the erosion identification process and the integration of texture and geometric allows us to obtain more homogeneous results, that could be improved by higher resolution images. The proposed approach is quite simple and can be easily integrated into current clinical workflows. In recent years, the trend of the research in the e-health field has moved towards the development of systems assembled as packages, associated with specific imaging modalities, such as CT, and MRI, and implemented as a part of PACS - Picture Archiving and Communication Systems~\cite{parascandolo2014computer}. The patient-specific characterisation of the vertebral spine permits the evaluation of the health status of each patient by analysing the spine in terms of anatomical components, functional regions and tissue status. The visualisation of the single bone and the whole district is improved, and the quantitative analysis relies on the information collected considering every direction in space, which is not possible with 2D images. According to the experimental validation, both the geometrical and the tissue analysis can help in the distinction between healthy and pathological subjects, supporting physicians in the evaluation of the subject's status.

The main limitation of our work is the need for a segmented 3D image since the 3D model is extracted from the segmented 3D image. Leveraging deep learning methods, the segmentation can be obtained automatically with good accuracy, thus reducing drastically the time required by this operation. Another limitation regards the subject's posture. In our study, the CT images show the patient in a supine position with an effect of gravity lower than the standing posture. The 2D CTs used in clinical practice have the great advantage to capture the standing patient, and thus they show the spine behaviour under different loads. However, all the methods developed are general enough to be applied to different imaging modalities and to further characterise the standing modality of the subjects. 

Future work will focus on improving the integration method toward a fully automatic approach, which would be easier in the presence of more than one follow-up exam, especially for a comparison that is considered extremely patient-specific. A higher number of exams, other than helping the automatic adjustment of the~$\epsilon$ parameter, could allow a more complete clinical validation. Another important aspect for further analysis is the spine characterisations while keeping a clinical application perspective, such as including an analysis of the intervertebral space and a further evaluation of the HU values inside the vertebral volume. Finally, the characterisation could be improved by considering further geometrical parameters and image pre-processing or enhancement techniques.

\paragraph{Acknowledgments}

This work has been supported by the European Commission, NextGenerationEU, Missione 4 Componente 2, ``\emph{Dalla ricerca all’impresa}'', Innovation Ecosystem RAISE ``\emph{Robotics and AI for Socio-economic Empowerment}'', ECS00000035.

\bibliographystyle{unsrtnat}
\bibliography{bibliography.bib}

\end{document}